\documentclass[manuscript,nonacm]{acmart} 

\AtBeginDocument{%
  }

\setcopyright{acmlicensed}
\copyrightyear{2026}
\acmYear{2026}
\acmDOI{}
\acmBooktitle{}

\begin{document}

\title{Why do we Trust Chatbots? From Normative Principles to Behavioral Drivers}

\author{Aditya Gulati}
\email{aditya@ellisalicante.org}
\orcid{0000-0002-0356-2987}
\affiliation{%
  \institution{ELLIS Alicante}
  \city{Alicante}
  \country{Spain}
}

\author{Nuria Oliver}
\email{nuria@ellisalicante.org}
\orcid{0000-0001-5985-691X}
\affiliation{%
  \institution{ELLIS Alicante}
  \city{Alicante}
  \country{Spain}
}

\begin{abstract}
  As chatbots increasingly blur the boundary between automated systems and human conversation, the foundations of trust in these systems warrant closer examination. While regulatory and policy frameworks tend to define trust in normative terms, the trust users place in chatbots often emerges from behavioral mechanisms. In many cases, this trust is not earned through demonstrated trustworthiness but is instead shaped by interactional design choices that leverage cognitive biases to influence user behavior. Based on this observation, we propose reframing chatbots not as companions or assistants, but as highly skilled salespeople whose objectives are determined by the deploying organization. We argue that the coexistence of competing notions of ``trust'' under a shared term obscures important distinctions between psychological trust formation and normative trustworthiness. Addressing this gap requires further research and stronger support mechanisms to help users appropriately calibrate trust in conversational AI systems.
\end{abstract}





\maketitle

\section{Introduction}

Trust is commonly defined as a psychological state involving the willingness of one party to accept vulnerability based on positive expectations about the intentions or behavior of another \cite{Rousseau1998}. In interpersonal contexts, trust emerges from perceptions of ability, benevolence, and integrity, and develops through repeated social interaction and moral accountability \cite{Mayer1995}. When extended to technological systems, trust shifts from intentions to performance expectations, reliability, and perceived control, giving rise to the concept of trust in automation \cite{Lee2004}. Importantly, trust does not necessarily require accurate understanding; individuals may trust under conditions of uncertainty by relying on cognitive heuristics and social cues rather than systematic evaluation \cite{Luhmann1979}. This distinction underpins the difference between \emph{how technology should be designed} and \emph{why people use it}.

Trust in traditional automation systems, such as those deployed in factories or industrial control environments however differs fundamentally from trust in conversational AI systems like chatbots. In industrial automation, trust is primarily grounded in observable system properties, including reliability, predictability, and performance consistency under well-defined operating conditions. Such trust is not spontaneous but is actively calibrated through formal training, standard operating procedures, certification processes, and clearly defined human oversight and intervention mechanisms \cite{Lee2004}. Operators are typically aware of the system’s functional scope and limitations, and trust develops gradually through repeated exposure and performance verification rather than through interactional cues.

Interaction patterns with automated systems, including how trust is formed and expressed, are also shaped by a range of cognitive biases, and particularly the \emph{automation bias}, \emph{i.e.}, the tendency of human operators to defer excessively to machine outputs, even in the presence of conflicting information. In industrial contexts, however, AI systems are rarely perceived as intentional or social agents. As a result, trust in industrial automation remains largely instrumental and task-oriented, grounded in expectations of mechanical correctness. 

Chatbots, by contrast, operate in domains traditionally governed by human judgment, advice, and social interaction, activating a different set of cognitive and social biases, such as the \emph{authority bias} that causes users to assume correctness because the system appears knowledgeable \cite{Milgram1963}; \emph{anthropomorphism} causing the attribution of human-like intentions or competence to chatbots \cite{Nass1994}; or the \emph{halo effect} generalizing from one perceived positive trait of the chatbot (\emph{e.g.}, fluency, politeness or aesthetic interface) to assume competence, accuracy or trustworthiness \cite{Nisbett1977,Gulati2024}. These factors illustrate that trust in chatbots is relational and psychologically constructed instead of procedural, and it is not necessarily evidence-based or morally warranted. As a result, users may believe the chatbot ``knows'' or ``cares'', rather than merely generates statistically plausible responses \cite{Langer1975,GliksonWoolley2020}.

Moreover, unlike face-to-face interactions or video-mediated communication, text-based chat conversations remove visual cues such as facial expressions, gaze, or perceived judgment. This absence creates a psychological sense of safety, reducing social anxiety and self-consciousness, which can encourage users to share information more freely and trust the system more readily. Interestingly, while companies could enhance chatbot interfaces with avatars or anthropomorphic faces, many seem reluctant to do so. One plausible reason is that adding a visible face or eyes may reduce the perceived anonymity and social distance, thereby making users less willing to disclose personal thoughts or accept guidance, which could paradoxically lower engagement and trust. In other words, the very ``invisibility'' of text-based chat interfaces acts as a cognitive shortcut, amplifying relational trust without the system actually providing moral or social accountability, and exploiting biases linked to comfort, perceived safety, and self-disclosure.

The existence of these cognitive biases in human-chatbot interaction makes them uniquely powerful and uniquely risky: trust is not earned through demonstrated competence alone but is cognitively constructed through conversational form, which can mask uncertainty, errors, and misalignment. As a result, trust in chatbots is often higher, faster, and less calibrated than trust in industrial automation, underscoring why traditional models of trust in automation may be insufficient for understanding human-AI interaction in conversational contexts.

\section{Rethinking the EU's pillars of trustworthy AI}

The development of trust in chatbots also diverges from the principles that policymakers and ethicists define as ``trustworthy AI'': it is frequently not the result of informed judgment but of cognitive shortcuts that reduce uncertainty and cognitive load. This phenomenon stands in tension with the European Commission’s seven requirements or pillars of trustworthy AI \cite{EUTrustworthyAI2019}, formulated prior to the widespread adoption of highly interactive conversational AI systems: human agency and oversight, technical robustness and safety, privacy and data governance, transparency, diversity and non-discrimination, societal and environmental well-being, and accountability. Rather than aiming to induce interpersonal trust in an AI system as a social counterpart, these requirements are meant to support lawful, ethical, and robust AI by enabling governance, auditability, contestability, and accountability, thereby promoting warranted reliance instead of passive acceptance.


Interestingly, greater alignment with these seven pillars could actually undermine rather than strengthen user trust in chatbots. Transparency, for example, may reveal opaque training processes, probabilistic reasoning, or the reuse of vast amounts of personal and public data, potentially triggering privacy concerns and skepticism. Similarly, explicit disclosures about limitations, biases, or error rates, central to technical robustness and accountability, can reduce the illusion of competence that users often project onto chatbots. From a cognitive perspective, ignorance can sustain trust: when users are unaware of data extraction practices or algorithmic uncertainty, they may feel more comfortable interacting with AI systems. Thus, while the pillars aim to promote ethically grounded and socially sustainable AI, they may clash with users’ preference for frictionless, seemingly authoritative tools, exposing a gap between normative trustworthiness and psychologically experienced trust.

In fact, empirical and cognitive research suggests that users often trust chatbots that would fail to meet several, if not most, of these pillars. Users readily over-trust chatbots based on linguistic fluency, social cues, and perceived competence, even when the systems operate opaquely, recycle personal or public data, or provide outputs that are only probabilistically plausible \cite{GliksonWoolley2020, Weizenbaum1966}. While the EU pillars remain vital for normative, rights-respecting design, they do not predict or explain actual human trust in chatbots. Thus, a reorientation may be needed: one framework for behavioral trust (why users trust) and one for normative trustworthiness (how systems should be designed).

\section{Pillars that drive trust in chatbots}

If trust in chatbots is treated as a behavioral and cognitive phenomenon rather than a normative ideal, a different set of ``pillars'' emerges, grounded in how users actually form trust judgments during interaction. Empirical and theoretical work in human-computer interaction and cognitive psychology suggests that user trust is driven less by transparency, accountability, or fairness, and more by surface-level interactional cues and experiential consistency. These trust-generating pillars are neither inherently ethical, nor are they aligned with the EU framework, but they exert powerful influence on user perception and behavior \cite{GliksonWoolley2020,Weizenbaum1966}.

\emph{First}, linguistic fluency and coherence function as primary trust heuristics. Due to well-documented cognitive biases, users tend to equate grammatical correctness, contextual relevance, and conversational smoothness with intelligence and reliability, even when underlying reasoning is shallow or probabilistic \cite{chiesurin2023,brunswicker2025}. \emph{Second}, perceived social presence and anthropomorphic cues play a central role in sustaining trust. Chatbots that adopt human-like language, express empathy, or simulate understanding elicit social responses that bypass critical evaluation. This dynamic echoes early findings from ELIZA onward, showing that users readily attribute intention, authority, and credibility to systems that merely appear conversationally competent \cite{Weizenbaum1966,lankton2015,klein2025,cao2025}. \emph{Third}, interactional consistency and responsiveness reinforce trust over time. When a chatbot provides timely, confident, and internally consistent responses, users develop expectations of reliability, regardless of whether the system is verifiable, unbiased, or transparent \cite{ali2025,vidarshika2025}. \emph{Finally}, frictionless usability and perceived instrumental value contribute to build trust due to pragmatic success. Chatbots that reduce effort, deliver answers quickly and integrate smoothly into workflows are trusted because they ``work", even if users lack insight into how or why they do so. Trust in this case is less of a moral judgment than a functional decision, arising from the convenience of use and outcome satisfaction \cite{vidarshika2025understanding}. 

\section{A sales person in every pocket}

While cognitive biases help explain why people trust chatbots, it is crucial to recognize that this trust does not necessarily reflect an alignment with the users’ interests. A useful reframing would be to treat chatbots not as companions, helpers, or assistants, but rather as highly skilled salespeople whose objectives are defined by the deploying organization and are rarely aligned with the well-being of their users. Unlike human assistants, chatbots are programmed to optimize engagement, collect data, or nudge users toward certain actions and goals that may conflict with the individual’s best interests. This tension is becoming increasingly salient in light of recent announcements that advertising may be incorporated directly into the outputs of widely used chatbots such as ChatGPT\footnote{\url{https://openai.com/index/our-approach-to-advertising-and-expanding-access/}}.

Research on effective sales strategies identifies key traits that build trust and facilitate persuasion, including empathy and active listening, persuasive and adaptive communication, deep product knowledge, integrity, and adaptability \cite{amor2019skills,shrestha2025sales}. These characteristics allow salespeople to make customers feel understood, reduce psychological resistance, and guide decision-making. Chatbots lack genuine empathy or ethical intent, yet they simulate many of these mechanisms through conversational design: natural language fluency, context-sensitive responses, politeness strategies, and apparent attentiveness mimic the social cues of a skilled salesperson. By doing so, chatbots exploit the same cognitive and emotional levers that make humans susceptible to influence, building trust without actual alignment to the user’s interests. This parallelism emphasizes why the metaphor of the chatbot as a ``trained salesperson'' is appropriate: like a highly skilled sellers, the system is designed to elicit trust and compliance, but unlike a human, it lacks moral accountability, highlighting the ethical and regulatory challenges of conversational AI.

This reframing has both theoretical and practical implications. Psychologically, it highlights that trust in chatbots is manufactured by design features that exploit human cognitive and social heuristics, rather than earned through reliability or moral alignment. Ethically, it underscores the importance of designing safeguards, transparency measures, and accountability mechanisms, treating chatbots more like persuasive agents in regulated contexts than genuine social actors. By acknowledging that chatbots function as strategic actors---much like skilled salespeople--- researchers, designers, and policymakers can better anticipate misuse, over-reliance, and privacy risks, while still supporting beneficial interactions in ways that respect users’ autonomy and rights.

\section{Discussion and Conclusion}
The analysis above suggests a dual framework for understanding trust in chatbots: one that differentiates why people trust them from how they should be designed ethically. Cognitive and social factors, including anthropomorphism, confirmation bias, sycophancy, and textual anonymity, create a powerful yet misleading sense of trust. Reframing chatbots as skilled salespeople rather than companions or helpers emphasizes that trust can be elicited through design features without guaranteeing alignment with users’ best interests. This perspective has several implications.

First, from a design perspective, developers should account for cognitive biases to calibrate trust appropriately, providing transparency, error signals, and corrective feedback while avoiding design choices that leverage cognitive biases to manipulate humans. Chatbots should also be designed to support user autonomy, for example by prompting reflection, offering alternative suggestions, and highlighting the limits of the system’s competence, helping users avoid over-reliance or inappropriate disclosure of sensitive information.

Second, from a regulatory and policy perspective, policymakers should distinguish between psychologically induced trust and normative compliance, ensuring that regulatory frameworks protect users even when AI systems are highly persuasive, and holding organizations accountable for misaligned outcomes. Context-sensitive oversight is critical, particularly in high-stakes domains such as healthcare, finance, legal advice, or education, where misaligned recommendations could have serious consequences. Organizations deploying AI systems should implement ethical auditing and accountability mechanisms, monitoring outcomes, fairness, and potential harms, while also providing clear disclosure of commercial intent and incentives so users understand whether the system is promoting engagement, collecting data, or advancing other organizational goals.

Third, from a societal standpoint, public awareness campaigns and interface design should communicate the limits of chatbot competence, helping users make informed decisions. Ensuring privacy and data protection is essential, including transparent data usage policies, minimal data retention, and robust consent mechanisms, given that chatbots often collect behavioral data to optimize engagement.

Finally, there is a need for empirical research to quantify the gap between perceived trustworthiness and actual alignment with ethical or normative AI standards, especially as generative AI systems become increasingly pervasive and influential. Longitudinal research should also examine the potential long-term effects on human decision-making, critical thinking, and behavior, as repeated interaction with persuasive systems may shape habits and cognitive strategies.

\begin{acks}
A.G. and N.O. are partially supported by a nominal grant received at the ELLIS Unit Alicante Foundation from the Regional Government of Valencia in Spain (Resolución de la Conselleria de Industria, Turismo, Innovación y Comercio, Dirección General de Innovación), along with grants from Intel and the European Union’s Horizon Europe research and innovation programme (ELLIOT; grant agreement 101214398). A.G. is additionally partially supported by a grant from the Banc Sabadell Foundation. This work was funded by the European Union. Views and opinions expressed are however those of the author(s) only and do not necessarily reflect those of the European Union or the European Commission. Neither the European Union nor the European Commission can be held responsible for them.

\end{acks}

\bibliographystyle{ACM-Reference-Format}
\bibliography{biblio}


\end{document}